%% file: acl_latex.tex
\pdfoutput=1
\documentclass[11pt]{article}
\usepackage{hyperref}
\usepackage{booktabs}
\usepackage{amsmath}
\usepackage{CJKutf8}
\usepackage{multirow}
\usepackage{graphicx}
\usepackage{booktabs}
\usepackage[]{acl}

\usepackage{caption}
\usepackage{subfigure}
\usepackage{color}

\usepackage{times}
\usepackage{latexsym}
\usepackage{verbatim} % group comment
\usepackage{float} % chart positon

\usepackage[T1]{fontenc}
\usepackage[utf8]{inputenc}

\usepackage{microtype}

\title{Statistical Dataset Evaluation: Reliability, Difficulty, and Validity}
\author{
Chengwen Wang\textsuperscript{\rm1 $*$}, Qingxiu Dong\textsuperscript{\rm1 $*$}, Xiaochen Wang\textsuperscript{\rm1}, Haitao Wang\textsuperscript{\rm2} and Zhifang Sui\textsuperscript{\rm1} \\
\textsuperscript{\rm 1} MOE Key Lab of Computational Linguistics, School of Computer Science, Peking University \\
\textsuperscript{\rm 2} China National Institute of Standardization  \\ 
  \texttt{\{wangcw,szf\}@pku.edu.cn, \{dqx,wangxiaochen\}@stu.pku.edu.cn
}\\
  \texttt{ 
  wanght@cnis.ac.cn
}
}

\begin{document}
\begin{CJK*}{UTF8}{gbsn}
\maketitle
\renewcommand{\thefootnote}{\fnsymbol{footnote}}
\footnotetext[1]{Equal contribution.}
\renewcommand{\thefootnote}{\arabic{footnote}}
\begin{abstract}
Datasets serve as crucial training resources and model performance trackers.
% in the Natural Language Processing (NLP) community. 
However, existing datasets have exposed a plethora of problems, inducing biased models and unreliable evaluation results.
In this paper, we propose a model-agnostic dataset evaluation framework for automatic dataset quality evaluation. 
We seek the statistical properties of the datasets and address three fundamental dimensions: reliability, difficulty, and validity, following a classical testing theory. 
Taking the Named Entity Recognition (NER) datasets as a case study, we introduce $9$ statistical metrics for a statistical dataset evaluation framework. 
% The metrics are well-formulated and show high agreement with manual evaluation
% we introduce $9$ statistical evaluation metrics for dataset quality evaluation and study how they affect the model performance.
%We find that Resume and CoNLL03 outperform other NER datasets in reliability and validity aspects, respectively, while for the difficulty, the WNUT16 dataset shows better differentiation and difficulty. 
% The assessment and conclusion manifest good agreement with human evaluation. And then, we reveal how the intrinsic properties reflect dataset quality and further influence model performance.  
Experimental results and human evaluation validate that our evaluation framework effectively assesses various aspects of the dataset quality.
Furthermore, we study how the dataset scores on our statistical metrics affect the model performance, 
%, and the proposed metrics are strongly correlated with model performance. 
%By data augmentation, we improve the model performance from XXX to X.% 
and appeal for dataset quality evaluation or targeted dataset improvement before training or testing models.\footnote{The code and data are available at \url{https://github.com/dqxiu/DataEval}.}

\end{abstract}

\section{Introduction}
Recently, a large number of models have made breakthroughs in various datasets of natural language processing (NLP)~\cite{bert,roberta}. Meanwhile, an increasing number and variety of NLP datasets are proposed for model training and evaluation~\cite{malmasi2022multiconer,yin2021docnli,srivastava2022beyond}.

\begin{figure}[htbp]
    %\centering
    \centerline{\includegraphics[width=1\linewidth]{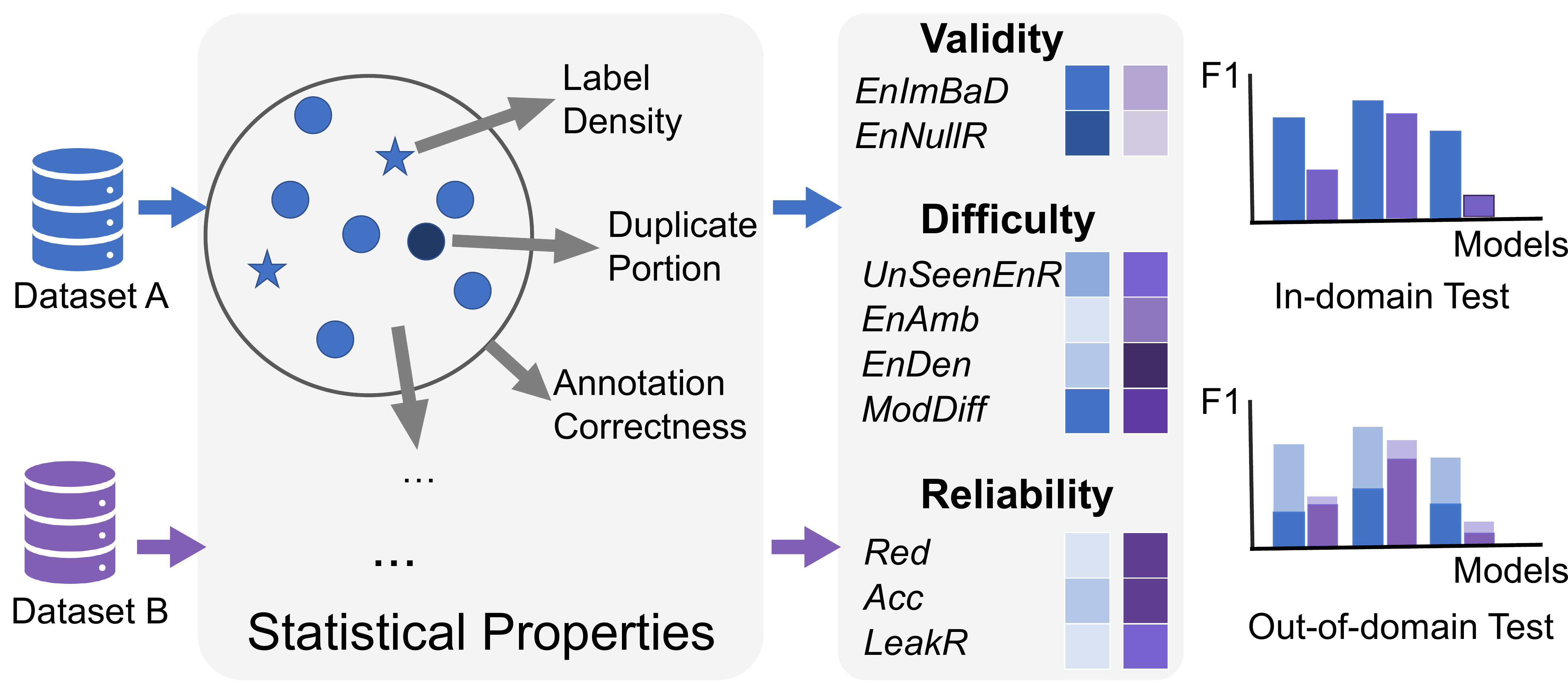}}
    \caption{Our statistical dataset evaluation framework based on the classical testing theory. We introduce $9$ quality evaluation metrics from three dimensions: reliability, difficulty, and validity. The dataset scores on the metrics exhibit a strong correlation with the models \footnote{Models trained on this dataset.} in many aspects, such as the average performance and the out-of-domain robustness.}
    \label{figure:1}
    \vspace{-0.2cm} %图片与正文间距
\end{figure}

However, despite the fact that datasets have a significant impact on model development and assessment~\cite{bommasani2021opportunities}, their quality is seldom systematically verified. 
Recent literature has indicated various quality issues within NLP datasets, e.g., label mistakes~\cite{wang-etal-2019-crossweigh}.
Datasets with quality issues frequently give rise to model shortcuts~\cite{gururangan-etal-2018-annotation,poliak-etal-2018-hypothesis} or induce incorrect conclusions~\cite{Goyal2022NewsSA,Rashkin2021MeasuringAI}.
% For example, \citet{wang-etal-2019-crossweigh} found 5.38\%  label mistakes in the CoNLL03 NER dataset. 
% Standard datasets in natural language inference (~\cite{gururangan-etal-2018-annotation};~\cite{poliak-etal-2018-hypothesis}) contain annotation artifacts that frequently give rise to spurious correlations or model shortcuts.  
% Datasets with quality issues may not effectively evaluate and differentiate the actual capabilities of different models or even induce incorrect conclusions~\cite{Goyal2022NewsSA,Rashkin2021MeasuringAI}.

In this paper, we aim to answer two primary questions: (1) How to evaluate dataset quality in a model-agnostic manner? A comprehensive dataset quality evaluation is crucial for selecting adequate training resources. Besides, when model performance conflicts across different datasets, the model-agnostic dataset quality evaluation results function as a good referee.
(2) How do the statistical scores on dataset properties affect the model performance? The answer will guide us in improving dataset quality for deriving effective and unbiased models.

To this end, we present a dataset evaluation framework through the statistical properties of the datasets (shown in Fig.~\ref{figure:1}).
Inspired by Classical Testing Theory (CTT)~\cite{novick1966axioms} in psychometrics, our dataset evaluation framework includes three key dimensions: reliability, difficulty, and validity. 
Reliability reflects how credible the dataset is, difficulty represents dataset difficulty and differentiation for models, and validity means how well the dataset fits the motivation and goal of the task.
In particular, we take the Named Entity Recognition (NER) datasets for a case study.
%, as NER is one of the most commonly researched and fundamental tasks in NLP.
We introduce $9$ metrics under the three dimensions for the statistical properties of NER datasets and assess the quality of $10$ widely used NER
datasets.
%the dataset evaluation in three aspects: reliability, difficulty, and validity. 
% Inspired by Classical Testing Theory (CTT) in psychometrics, we propose to evaluate the dataset quality from three dimensions: reliability, difficulty, and validity. 
%Reliability includes 3 statistical metrics, which reflects how credible the test results of the dataset are. Difficulty includes 4 statistical metrics, representing the degree of difficulty and differentiation of the datasets for the existing model. Validity includes 2 statistical metrics and means how well the evaluation dataset fits the motivation and goal of the evaluation.
% All the metrics are derived from statistical properties for the NER tasks and are validated by human evaluation.
% We performed dataset quality evaluation on 10 commonly used NER datasets and implemented experiments to to verify the validity of our proposed metrics assessing the quality of the dataset. 
%[Add technical details that reflect the challenging nature of our research questions and the novelty of our approach]

Extensive experimental results validate that our evaluation metrics derived from the dataset properties are highly related to the performance of NER models and human evaluation results. 
The evaluation results enhance our comprehension of the datasets and bring some novel insights. For example, one of the most widely used English NER datasets, \textit{CoNLL03}~\cite{sang2003introduction}, is far less challenging ($0.43$, $0.30$, and $2.63$ points lower on the Unseen Entity Ratio, Entity Ambiguity Degree and Model Differentiation metrics, respectively) than \textit{WNUT16} \cite{strauss-etal-2016-results}, which has received less attention previously.
In addition, by controlled dataset adjustment ~(Sec.~\ref{sec:63}), we find the dataset quality on the statistical metrics, including %Leakage Ratio, 
Unseen Entity Ratio, Entity Ambiguity Degree, and En-Null Rate deriving affects the NER model performance significantly. Therefore, we appeal for dataset quality evaluation or targeted dataset improvement before training or testing models.

We believe that statistical dataset evaluation provides a direct and comprehensive reflection of the dataset quality. And we recommend dataset quality evaluation 
 before training or testing models for a better understanding of tasks and data for other tasks in NLP.

\section{Classical Testing Theory}
Human tests or exams usually follow strict testing theories, such as Classical Testing Theory (CTT)~\cite{novick1966axioms}, a statistical framework to measure the quality of the exams. CTT considers that a comprehensive and systematic evaluation should consider three dimensions: reliability, difficulty, and validity.

In this paper, we introduce \textbf{CTT for Dataset Evaluation}. Adapting traditional CTT to dataset evaluation, we specified the definitions of reliability, difficulty, and validity as follows:

\begin{itemize}
    \item \textbf{reliability}~ measures the reliability of the evaluation dataset. For instance, those datasets with many labeling errors do not have sufficient confidence to evaluate the performance of different models.

    \item \textbf{Difficulty}
~is used to evaluate the difficulty of the dataset and its differentiation between different models and man-machine performance.

    \item \textbf{Validity}~ aims to evaluate how well the dataset effectively measures the capability of models.

\end{itemize}

\section{Dataset Quality Evaluation Framework}
\label{sec:3}

Following CTT for Dataset Evaluation, we build our statistical dataset evaluation framework and apply it to NER datasets. It includes nine fundamental metrics of the statistic properties in the NER datasets. In this section, we introduce the definition and the mathematical formulations of the proposed metrics. 

For a dataset\footnote{Usually, the dataset includes the training set, the development set, and the test set.} $D$ with $n$ instances, let $(x^{(i)}, y^{(i)})$ represent the i-th instance ($i=1,2,...,n$) . The input sequence $x^{(i)}$ consists of $m^{(i)}$ tokens, and the output sequence $y^{(i)}$ consists of $m^{(i)}$ entity labels. Let $\mathcal{C}$ represent the entity types in $D$ (including ``Not an entity'' ), and each entity type $c_j \in \mathcal{C}, j \in {1,2,...,v}$, where $v$ represents the total number of entity types. We use $Te, Tr, De$ to represent the test set, the training set, and the development set, respectively. The function $e(\boldsymbol{y_D})$ is defined to obtain a set of entities in the set of $y^{(i)}$ of $D$, $\boldsymbol{y}$, and sometimes we omit $D$ for simplification.

\input{floats/r-d-v-result.tex}

\subsection{Metrics under Reliability}\label{formu}

The metrics under reliability aim to evaluate how accurate and trustworthy a dataset is, including Redundancy, Accuracy, and Leakage Ratio.

\paragraph {Redundancy} represents the proportion of duplicate instances within a dataset~$D$.   
\begin{equation*}
\text{Red}(D) = \frac{\sum_{i=1}^{n} \sum_{j=i+1}^{n} [(x^{(i)}, y^{(i)}) = (x^{(j)}, y^{(j)})]}{n}
\end{equation*}

To calculate the Redundancy of a dataset, we first count the number of the same instances that appeared more than once in the dataset, then use it to divide the total number of instances in the dataset.

\paragraph {Accuracy} aims to evaluate the annotation correctness of the dataset and can be calculated as follows:
\begin{align*}
\delta(x^{(i)}, y^{(i)}) = \begin{cases}
1, & \text{if } y^{(i)} \text{~is accurate for~} x^{(i)}, \\
0, & \text{else} 
\end{cases}
\end{align*}
\begin{equation*}
\text{Acc}(D) = \frac{\sum_{i=1}^{n} \delta(x^{(i)}, y^{(i)})}{n}
\end{equation*}

We recommend selecting 100 instances from each dataset split and inviting at least $3$ professional linguists to annotate the accuracy. 
% When the annotator thinks that all entity boundaries and types in an instance are accurately labeled, it is marked as true, otherwise it is false. 
The value of Cohen Kappa \cite{cohen1960coefficient} between the annotators should be higher than $0.75$.

\paragraph {Leakage Ratio} is used to measure how many instances in the test set have appeared in the training set or development set, defined as:
\begin{equation*}
\text{LeakR}(D)\!=\!\frac{\sum_{i=1}^{|Te|} [(Te^{(i)} \in Tr) \text{ or } (Te{(i)} \in De)]}{n}
\end{equation*}

%We count the number of instances in the test set that appears in the training set or the development set first and divide it by the number of test set instances to get the Leakage Ratio of one particular NER dataset.

\subsection{Metrics under Difficulty}
We propose $4$ metrics under difficulty to assess how challenging the datasets are, including three intrinsic metrics (Unseen Entity Ratio, Entity Ambiguity Degree, and Text Complexity) and one extrinsic metric (Model Differentiation).

\paragraph {Unseen Entity Ratio}  
We use the Unseen Entity Ratio to measure the proportion of entities in the labels of the test set that have never been presented in the training set. It is calculated as:
     \begin{equation*}
     \text{UnSeenEnR}(D) = \frac{|e\left(\boldsymbol{y_{Te}}\right) \setminus 
{e(\boldsymbol{y_{Tr}})}|}{|e(\boldsymbol{y_{Te}})|}
    \end{equation*}

\paragraph{Entity Ambiguity Degree} is mainly used to measure how many entities are labeled with more than one kind of entities types. For example, if ``apple'' is labeled as ``Fruit'' in one instance and labeled as ``Company'' in another instance, then there is a conflict in $D$. We introduce $e^*(D)$ to represent the number of conflict entities in dataset $D$ and obtain the Entity Ambiguity Degree by:
    \begin{equation*}
     \text{EnAmb}(D) = 1 -\frac{e^*(D)}{n} 
    \end{equation*}

\paragraph{Text Complexity} means the entity density of sentences in $D$ on average. It is simply formulated as follows:
\begin{align*}
    \text{EnDen}(D) = \sum_{i=1}^n \frac{|e\left(y^{(i)}\right)|}{n m^{(i)}}
\end{align*}

\paragraph{Model Differentiation} is used to measure how well a dataset distinguishes the performance of different models. Here, we use the standard deviation of the scores of $k$ different models trained on the same dataset to indicate how well the dataset distinguishes between different models.  
    \begin{align*}
       \text{ModDiff}(D) =  \text{Std}(\theta_1,\theta_2,\cdots,\theta_k)
    \end{align*}    
%where Std(·) is the function to compute the standard deviation. 
We recommend using the top 5 model scores on the dataset\footnote{From the Paperswithcode website.} for ModDiff calculation. 

\subsection{Metrics under Validity}

The metrics under validity, for example, Entity Imbalance Degree and Entity-Null Rate for NER datasets, are mainly proposed to evaluate the effectiveness of the dataset in evaluating the model's ability on the specific task.

\paragraph{Entity Imbalance Degree} mainly measures the unevenness of the distribution of different entities in $D$. Specifically, we use standard deviation to quantify the degree of dispersion of the distribution of all the different types of entities $\mathcal{C}$ in the dataset.  
    \begin{equation*}
    \fontsize{10pt}{\baselineskip}\selectfont
    \begin{split}
    \text{EnImBaD}(\mathcal{D}) \!=   \!Std\left(P_{y_D}(c_1),P_{y_D}(c_2),\cdots\!,P_{y_D}(c_v)\right)
    \end{split}
    \end{equation*}

\paragraph{Entity-Null Rate} measures the proportion of instances that do not cover any entity and is defined as follows:
\begin{align*}
\zeta(y^{(i)}) = \begin{cases}
1, & \text{if } y^{(i)} \text{~has no entity}, \\
0, & \text{else} 
\end{cases}
\end{align*}

\begin{equation*}
\text{EnNullR}(D) = \frac{\sum_{i=1}^{n} \zeta(y^{(i)})}{n}
\end{equation*}

% To obtain the Entity-Null Rate, we count the number of instances in the set that does not contain any entities and divide it by the overall instance number.

\begin{figure}[]
\centering
\subfigure[]{
\begin{minipage}[t]{0.55\linewidth}
\centering
\includegraphics[width=\linewidth]{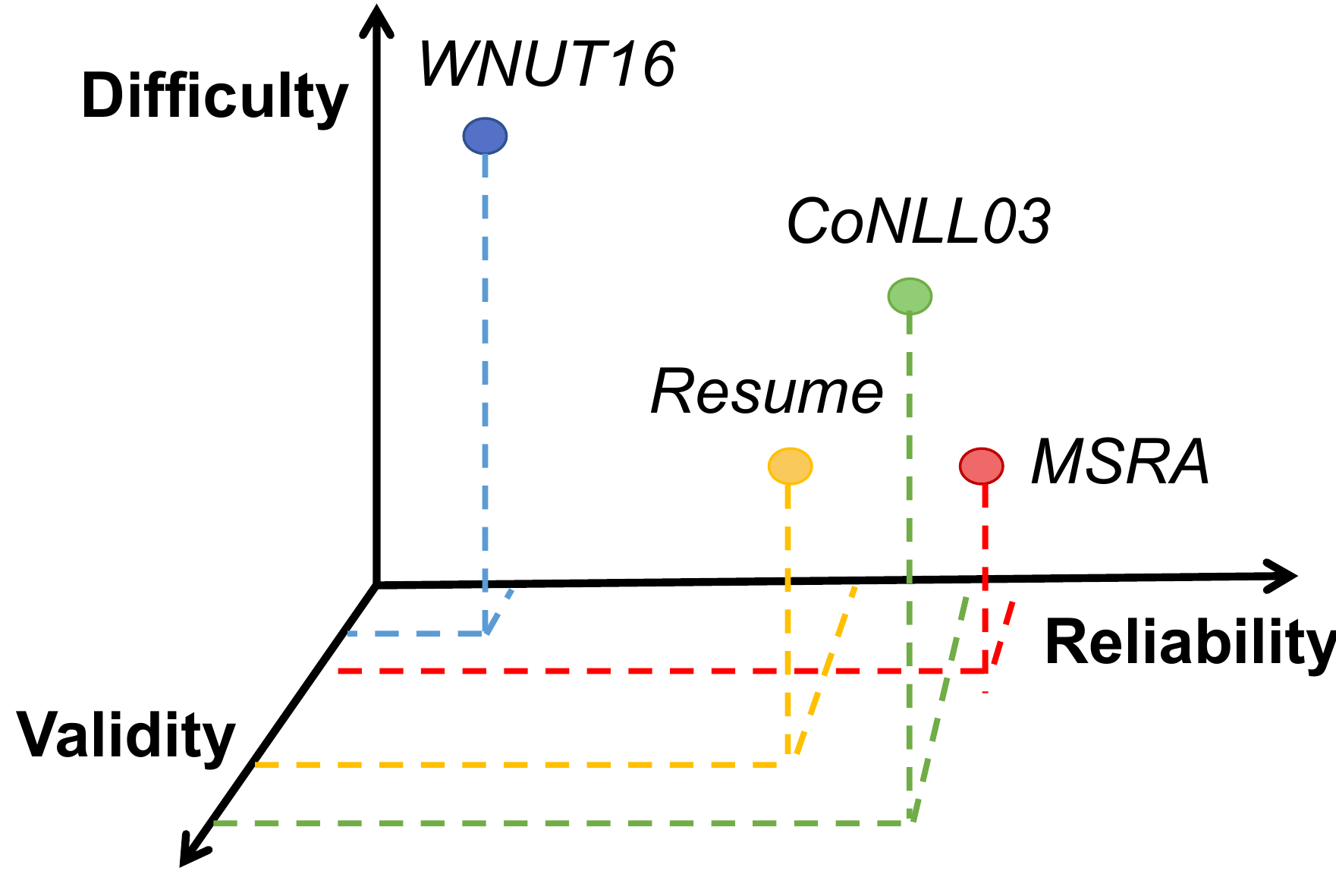}
%\caption{fig1}
\end{minipage}%
}%
\subfigure[]{
\begin{minipage}[t]{0.45\linewidth}
\centering
\includegraphics[width=\linewidth]{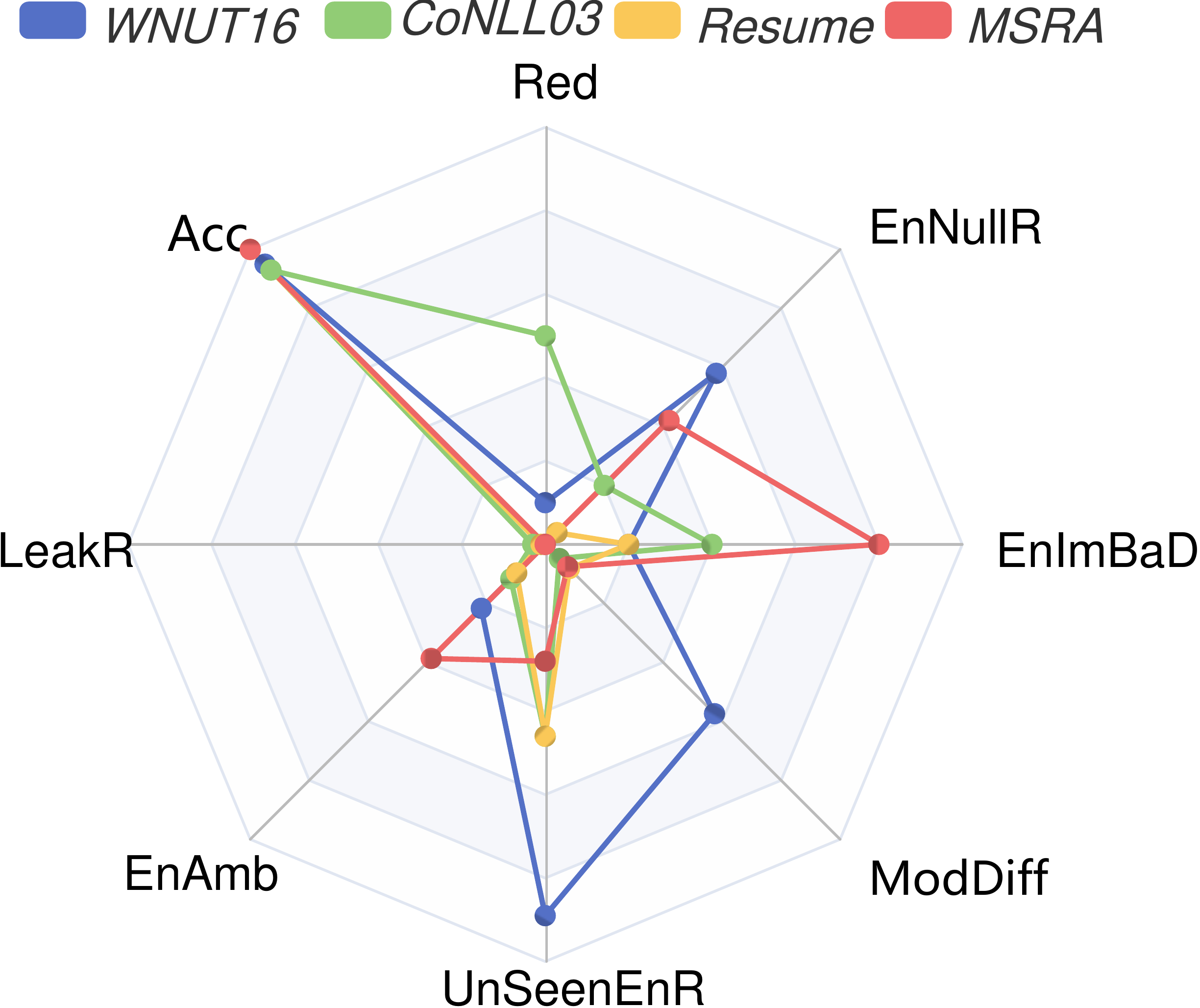}
%\caption{fig2}
\end{minipage}%
}%
\quad
\centering
\caption{Evaluation results of WNUT16, CoNLL03, Resume, MSRA under different dimensions and metrics. The abbreviations and corresponding full names of the metrics are presented in Sec.~\ref{sec:3}.}
\end{figure}

\section{Statistical Dataset Evaluation for NER}
To validate our statistical dataset evaluation methods, we assess the quality of $10$ widely used NER datasets, including $3$ English NER datasets and $7$ Chinese NER datasets. The evaluation results are shown in Tab.~\ref{tab:r-d-v-results}.

\subsection{Datasets}
We provide the basic information about the datasets in Tab.~\ref{tab:dataset info}. 

\paragraph{English NER Datasets}

\textit{CoNLL03 NER}~\cite{sang2003introduction} is a classical NER evaluation dataset consisting of 1,393 English news articles.
\textit{WNUT16 NER} \cite{strauss-etal-2016-results} is provided by the second shared task at
WNUT-2016 and consists of social media data from Twitter. 
% It includes 10 entity types.
\textit{OntoNotes5} \cite{weischedel2013ontonotes} is a multi-genre NER dataset collected from broadcast news, broadcast conversation, weblogs, and magazine genre, which is a widely cited English NER dataset.

\paragraph{Chinese NER Datasets}
\textit{CLUENER}  ~\cite{xu2020cluener2020} is a well-defined named entity recognition dataset with more and finer-grained entity types. Specifically, apart from standard labels like person, organization, and location, it contains more diverse categories such as Company, Game, and Book, etc.
% In our knowledge, CLUENER2020 is a CNER dataset with the most entity types, including a total of 10 entity categories.
% in Chinese which is introduced by  CLUE organization.
\textit{OntoNotes4} ~\cite{weischedel2011ontonotes} is copyrighted by Linguistic Data Consortium\footnote{https://www.ldc.upenn.edu/} (LDC), a large manual annotated database containing various fields with structural information and shallow semantics. 
% (news, blogs, magazines, telephone speech, etc.)
%Although 18 entity types are annotated in the dataset,  in practical Person, Location, Organization, and Geo-political are often used in CNER task.
\textit{MSRA} \cite{levow2006third} is a large NER dataset in the field of news, containing distinctive text structure characteristics.
%three categories of entities, namely Person, Location, and Organization, 
\textit{PeopleDaily NER}\footnote{https://github.com/zjy-ucas/ChineseNER} is a very classic benchmark dataset to evaluate different NER models. 
%As with most Chinese NER datasets, it includes three types of entities: Location, Organization, and Person.
\textit{Resume NER} ~\cite{zhang-yang-2018-chinese} is a resume dataset with high annotator agreement, which consists of resumes of senior executives from listed companies in the Chinese stock market. 
%They randomly select 1027 resume summaries and manually annotate 8 types of named entities with YEDDA system \cite{yang1711lightweight}. The inter-annotator agreement is 97.1\%.
\textit{Weibo NER}  \cite{peng-dredze-2015-named,he-sun-2017-f} is a dataset from the social media domain of Sina Weibo. 
% It contains four types of entities: Person, Location, Organization and GPE.
\textit{WikiAnn} \cite{pan-etal-2017-cross}
is a Chinese part of a multilingual named entity recognition dataset from \href{https://www.wikipedia.org/}{Wikipedia}
 articles.
 %  annotated with LOC (location), PER (person), and ORG (organisation) labels

\input{floats/dataset_info}

\subsection{Settings}
\label{sec:42}
According to the metrics we proposed in Sec.~\ref{sec:3}, we calculate the statistical scores for each dataset. Specifically, we average the scores of the training, the development, and the test split of the datasets for Redundancy, Accuracy, Entity Ambiguity Degree, Entity Density, Entity Imbalance Degree, and Entity-Null Rate, respectively. For Leakage Ratio, Unseen Entity Ratio, and Model Differentiation, we only calculate the scores on the specific splits involved according to Sec.~\ref{formu}.

\subsection{Dataset Reliability}

\subsubsection{Annotation Accuracy}
Accuracy scores quantitatively inform us that we cannot take it for granted that all benchmark datasets are reliable.

We observe that \textit{CLUENER} has the lowest accuracy score. In particular, it has 17\% errors in its development set~(shown in Appendix~Tab.~\ref{tab:16}). Conversely, the other datasets (e.g., \textit{Resume} and \textit{WNUT16}) have a relatively high accuracy score for both Chinese and English NER datasets. 

\subsubsection{Leakage Ratio}
The dataset's shortcomings (under the reliability dimension) can be effectively revealed by the Leakage Ratio. Given the Leakage Ratio results, we are surprised to find that \textit{Weibo} and \textit{WikiAnn}  have serious data leakage issues.

\begin{figure}[]
    %\centering
    \centerline{\includegraphics[width=1\linewidth]{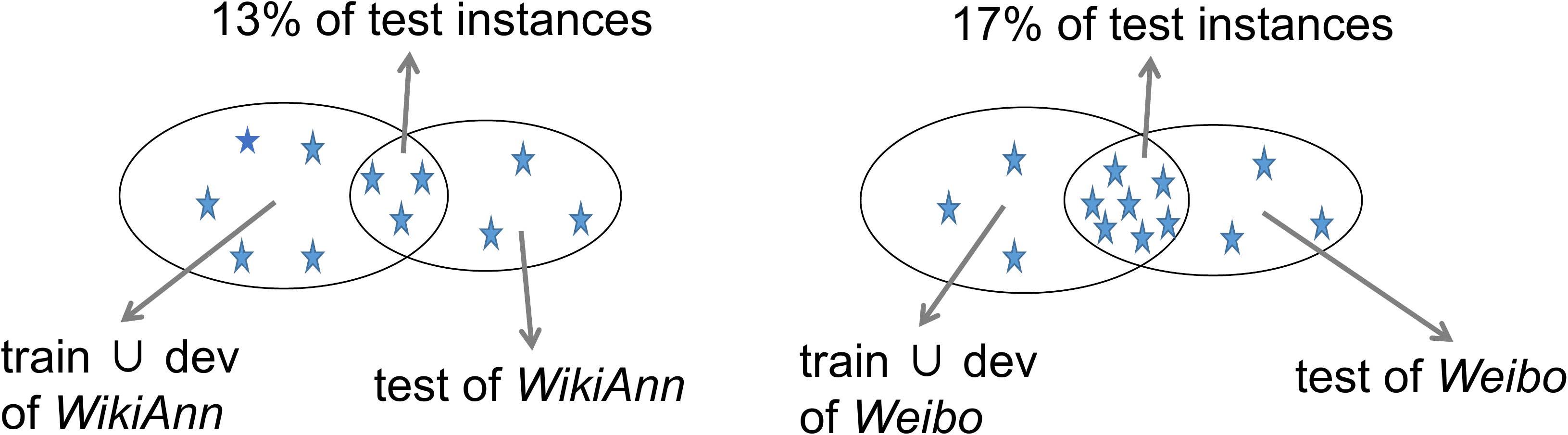}}
    \caption{Leakage Ratio values of \textit{WikiAnn} and \textit{Weibo}. It is observed that 0.13~(13\%) and 0.17~(17\%) of the instances in the test set of \textit{WikiAnn} and \textit{Weibo}, respectively, appear in their corresponding training or development sets.}
    \label{figure:Leakage-schematic diagram}
    \vspace{-0.2cm} %图片与正文间距
\end{figure}

As shown in Tab.~\ref{tab:r-d-v-results} and Fig.~\ref{figure:Leakage-schematic diagram}, 0.17~(17\%) and 0.13~(13\%) of the instances in the test set of \textit{Weibo} and \textit{WikiAnn} have appeared in their corresponding training or development sets, respectively. 
%Intuitively, deep learning models, especially large models, are more likely to perform better in the test phase on instances that have already been encountered in the training phase. We will validate this conclusion further in Sec.~\ref{exp-results}.

\subsubsection{Overall Reliability}

Combining several metrics under the reliability dimension in Tab.~\ref{tab:r-d-v-results}, we can conclude that \textit{Resume} and \textit{MSRA} maintain high reliability.

In specific, there is no data redundancy in \textit{Resume} and \textit{MSRA}. That is to say, the instances of each part of the dataset are unique and non-repeating. Additionally, they achieve the highest Accuracy scores and hardly show data leakage problems, with a Leakage Ratio of 0.00 and 0.01, respectively.

\subsection{Dataset Difficulty}

\subsubsection{Unseen Entity Ratio}

Results on Unseen Entity Ratio~(UnSeenEnR) demonstrate the generalization ability of NER models on unseen entities.

The evaluation results show that \textit{Weibo} and \textit{WNUT16} are more difficult in terms of UnSeenEnR because their test sets have a 0.56 and a 0.89 ratio of entities that have not appeared in training, respectively. \textit{WikiAnn} is the Chinese dataset only second to \textit{Weibo} that can better evaluate the generalization ability of NER Models.
Conversely, \textit{PeopleDaily NER} and \textit{OntoNotes5} are suboptimal for evaluating model generalization ability. Our experiment results in Sec.~\ref{exp-results} reveals that model trained on them are more likely to perform better on seen entities compared to those that have not appeared in the training set. 

\subsubsection{Entity Ambiguity Degree}
Entity Ambiguity Degree (EnAmb) captures observable variation in the information complexity of datasets.

Given our findings,
\textit{OntoNotes~4} and \textit{WNUT16} are the Chinese and English NER datasets with the highest Entity Ambiguity Degree, respectively, which means that they are more difficult for models to accurately predict entity types. Consistent with our conclusion (in Sec.~\ref{exp-results}), \citet{bernier2020hardeval} also argue that SOTA models cannot be able to deal well with those entities labeled with different types in different contexts.

\subsubsection{Model Differentiation}

Extrinsic evaluation metrics, such as Model Differentiation) are also necessary for evaluating the difficulty of datasets.

Unlike those intrinsic evaluation metrics (e.g., Entity Ambiguity Degree), \textbf{Model Discrimination} (ModDiff) aims to assess the dispersion of model scores on a unified benchmark dataset. That is to say, a more difficult dataset should have a clear distinction between models with different abilities. As shown in Tab.~\ref{tab:r-d-v-results}, \textit{CLUENER} and \textit{WNUT16} are Chinese and English datasets that can better distinguish model performance, respectively.

\subsubsection{Overall Difficulty}

\textit{WNUT16} is a more difficult benchmark for English NER as a whole.

Although \textit{WNUT16} has less citations than \textit{CoNLL03} and \textit{OntoNotes5},
as demonstrated in Tab.~\ref{tab:r-d-v-results}, \textit{WNUT16} has a higher Entity Ambiguity Degree and Unseen Entity Ratio than the other two English NER datasets. Meanwhile, we find that the model performance gap on \textit{WNUT16} is large, indicating that it is more difficult and can effectively distinguish models with different performances.

\subsection{Dataset Validity}
\subsubsection{Entity Imbalace Degree}

Datasets with uneven distribution of entity types may not effectively evaluate the ability of models on the long-tailed instances.

Intuitively, the model does not perform as well on those long-tailed entity types as other entities.
We observe that \textit{Weibo}  achieves the highest Entity Imbalance Degree (EnImBaD) by a large margin, indicating that its distribution of entity types is heavily uneven.
Therefore, datasets with severely uneven distribution of entity types can only evaluate the performance of the models on a large number of distributed entity types.

\subsubsection{Entity-Null Rate}

Surprisingly, there are a large number of instances without any entities in many datasets such as \textit{OntoNotes4}, \textit{MSRA}, \textit{WNUT16} and \textit{OntoNotes5}.

%NER datasets mainly focus on intensively testing the entity recognition capabilities of NER models. 
Although certain naturally distributed texts will contain some sentences without named entities, a high number of entity-free samples in a NER dataset makes it impossible to give a sufficient number of instances for NER model validation.

%Those without any entities do not provide a large number of entities for the evaluation of models. 
%We speculate that the existence of those examples without any entities is mainly due to the fact that in the dataset construction stage, the dataset constructor uses paragraphs as annotation objects, and not every sentence in these paragraphs has entities. 
%Generally in NER tasks，models use sentences as processing units and do not need to use paragraph structure information. 
%Therefore, we believe that dataset builders should pay attention to those sentences without any entities, and construct entity-rich datasets for evaluating the entity recognition ability of models. 

\subsubsection{Overall Validity}
In general, \textit{CoNLL03} is the English NER dataset with the highest validity.
As shown in Tab.~\ref{tab:r-d-v-results}, \textit{CoNLL03} has the lowest EnNullR, indicating that it can intensively test the entity recognition capabilities of NER models.

\input{floats/ennullr028.tex}

\input{floats/chnullr028.tex}

\section{How Dataset Properties Affect Model Performance?}
\label{sec:validate}
% We built the evaluation metrics for NER datasets in terms of dataset properties and conducted a quality assessment of 10 NER datasets.
To validate the metrics and results under our statistical evaluation framework and to further investigate how the statistical metric scores on dataset properties affect the model performance, we conduct controlled dataset adjustment in this section.

\subsection{Models}
We use $3$ models for experiments on the Chinese NER datasets, including:
1) \textbf{Lattice LSTM} ~\cite{zhang2018chinese}, a model based on the Long Short-term Memory Networks \cite{chiu2016named}, which finds the valuable words from the context through the gated recurrent cells automatically. 
% Besides, there is no word segmentation error. 
%So it performs better than both character-based and word-based methods.
 %The disadvantage is information loss and low computing performance.
2) \textbf{Flat-Lattice}~\cite{li2020flat}, which
converts the lattice structure into a flat structure 
%consisting of spans and introduce vocabulary information without loss by well-designed position encoding.
%It also supports parallel computing and improves the inference speed well.
3) \textbf{Roberta} ~\cite{roberta}, a transformer-based pretrained model which removes the next sentence predict task in BERT.
 % and changes the masking strategy from static to dynamic
 %with larger model parameters, data, batch\_size.
 
For the English datasets, we also take $3$ models, including:
1) \textbf{LSTM CRF} ~\cite{lample2016neural}, a traditional model based on the bidirectional LSTM with conditional random fields (CRF). 
%where CRF imposes strong constraints on the label of each word in tag sequences.
% LSTM has been designed and proved to learn long dependencies by incorporating a memory-cell. 
2) \textbf{LUKE} \cite{yamada2020luke}, which
provides new pretrained contextualized representations of words and entities by predicting masked words and entities in entity-annotated corpus based on the bidirectional transformer ~\cite{vaswani2017attention}. 
% LUKE introduces an entity-aware self-attention mechanism, which extends the Transformer. 
3) \textbf{W2NER}~\cite{li2022unified}, which
%is a unified model for both Chinese and English. It can also achieve good results on datasets with flat, overlapping, and discontinuous entities. Because the model 
converts NER to word-word relationship classification and models the neighboring relations between entity words with Next-Neighboring-Word (NNW) and Tail-Head-Word (THW) relations.

\subsection{Experiment Settings}

All the experiments are done on the NVIDIA RTX 2080 GPU and 3090 GPU and evaluated by seqeval~\footnote{https://github.com/chakki-works/seqeval}. For the experiment with Train-Dev dataset adjustment~(Sec.~\ref{sec:63}), we report the averaged results and variances over three random seeds.

\subsubsection{Hyper-parameters and Word Embeddings}

%Specifically, W2NER, Roberta and LUKE adopt AdamW \cite{loshchilov2017decoupled} optimizer. Stochastic Gradient Descent (SGD) is used for optimation in Lattice LSTM, BiLSTM-CRF, and Flat Lattice. 
%(move to the appendix)%LUKE and W2NER optimize the model using AdamW with learning rate warmup and linear decay of the learning rate. LUKE also uses early stopping based on performance on the development set.

Specifically, we choose the no-BERT version Flat Lattice and add a CRF layer on Roberta. We select the best performance version of LSTM CRF with pretrained word embeddings, character-based modeling of words, and dropout rate. The more detailed settings and hyper-parameter are described in Appendix~\ref{app:setting}.
As known in~\cite{lample2016neural}, models using pretrained word embeddings perform better than randomly initialized ones. Therefore, we use different word embedding methods to ensure the comprehensiveness of the experiment. The specific word embeddings selection is shown in Appendix~\ref{app:emb}. 

\begin{figure}[]
    %\centering
    \centerline{\includegraphics[width=1\linewidth]{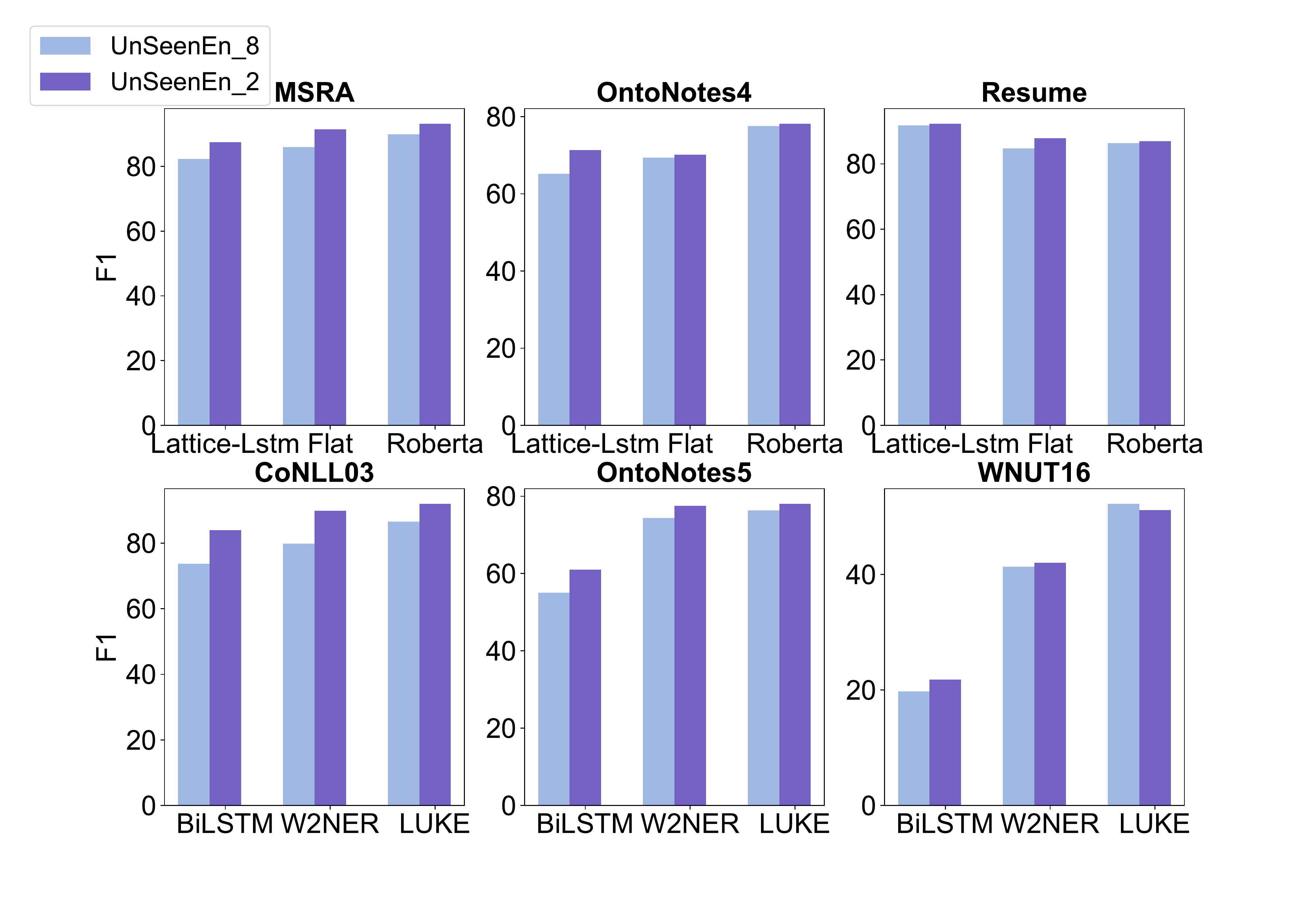}}
    \caption{Model performance on NER datasets when the proportion of unseen entities~(UnSeenEn) in the test set is 0.80~(80\%) and 0.20~(20\%), respectively.}
    \label{figure:3}
    \vspace{-0.2cm} %图片与正文间距
\end{figure}

\subsection{Controlled Dataset Adjustment}
\label{sec:63}

 We conduct controlled dataset adjustment experiments to investigate how the statistical properties (measured by our metrics) affect model performance: 1) modifies the test set to construct two new test sets (of the same size) of which statistical values of one certain metric (e.g., Leakage Ratio and Unseen Entity Ratio) are clearly distinguishable (i.e., Test Dataset Adjustment). 
2) modifies training set and development set to construct two new training sets and development sets (of the same size), respectively, of which statistical values of one certain metric are clearly distinguishable (i.e., Train-Dev Dataset Adjustment).

\subsubsection{Test Dataset Adjustment}

We adjust the test set around two metrics of %Leakage Ratio, 
Unseen Entity Ratio, and Entity Ambiguity Degree to construct two new test sets respectively, and make their statistical values on the corresponding metrics relatively discrete. For example, as for the Unseen Entity Ratio, we adjust the test set to construct two new test sets, one with an Unseen Entity Ratio of 0.80~(80\%) and the other with an Unseen Entity Ratio of 0.20~(20\%), while ensuring that the two newly constructed test sets have the same number of instances.

\subsubsection{Train-Dev Dataset Adjustment}

First, we selected some datasets with high En-Null Rates, i.e., \textit{WNUT16} and \textit{OntoNotes5} for English datasets, \textit{Weibo} and \textit{OntoNotes4} for Chinese datasets. Then we filter the data in the training set and development set to make the En-Null Rate 20\% and 80\%, respectively, and ensure that the instance numbers of these two subsets are consistent. Finally, we use different models to train the data and then use the same test set to test and compare the scores.

\subsection{Experiment Results and Analysis}

\label{exp-results}

% \paragraph{The model performs better on the instances it has seen during training.}
% As shown in Tab.~\ref{tab:LeakageExp}, 3 models (i.e., Lattice LSTM, Flat-Lattice and Roberta) consistently achieve better performance when the leakage rate of the test set is 80\% than when it is 20\%. In particular, we found that the performance of  Flat-Lattice on Weibo test set with a Leakage Ratio of 80\% outperformed the 20\% by a large margin i.e. 25.69\%. We speculate that because the model has seen the leaked data in the test set during training, it performs better on the test set with a relatively high data leakage rate. Looking at the experimental results from another perspective, researchers need to pay more attention to how to improve the generalization ability of the NER model.

\paragraph{Datasets with high Unseen Entity Ratio are more difficult for NER models.}

Intuitively, those entities that were seen during training are more challenging for NER models compared to those that did not appear in the training set. Fig.~\ref{figure:3} supports our intuition. Models perform better on datasets with a lower proportion of unseen entities than in datasets with a relatively high proportion of unseen entities.

\paragraph{Entities with strong Entity Ambiguity Degree are indeed more likely to confuse the model.}

\begin{figure}[]
    %\centering
    \centerline{\includegraphics[width=1\linewidth]{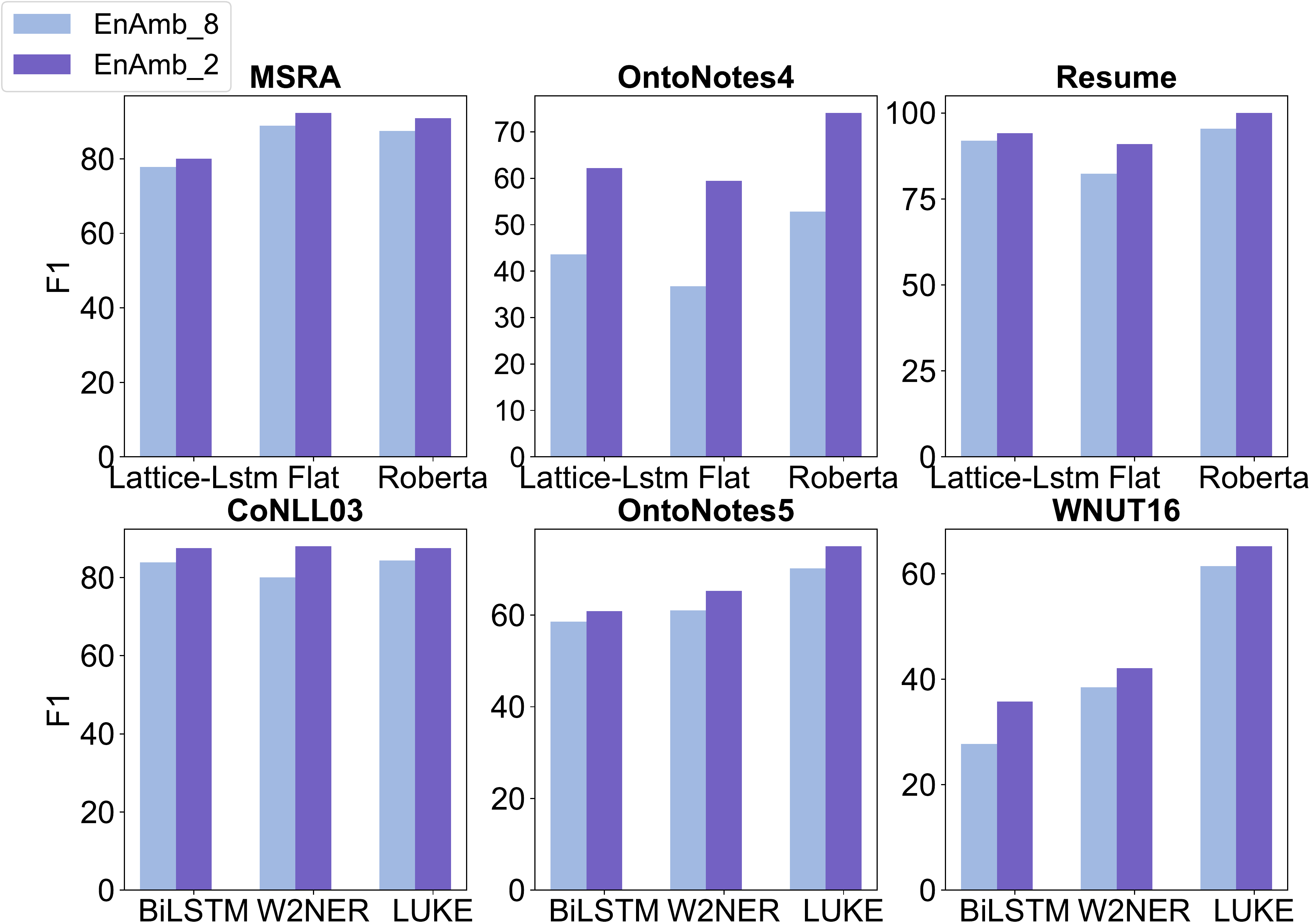}}
    \caption{Model performance on NER datasets when the proportion of ambiguous entities in the test set is 0.80~(80\%) and 0.20~(20\%), respectively.}
    \label{figure:4}
    \vspace{-0.2cm} %图片与正文间距
\end{figure}

We can infer from Fig.~\ref{figure:4} that datasets with a high Entity Ambiguity Degree are more challenging for the model. As for models tested on Chinese datasets, their average performance is 6.42 (F1) points higher on datasets with high entity ambiguity rates than on datasets with low entity ambiguity rates. The English NER model is more likely to be confused by entities with a high entity ambiguity rate and make wrong decisions.

\paragraph{Entity-Null Rate plays a small difference.}
As shown in Tab.~\ref{tab:ennullr028} and Tab.~\ref{tab:chnullr028}, the F1 score of the training set and development set with EnNullR of 0.20~(20\%) is better than 0.80~(80\%). Therefore, we conclude that the contribution of instances without entities to the model is less than the instances with entities during training. However, are instances without any entities completely useless for model training? We delete all these instances and show the results in Tab.~\ref{tab:ennullr028} and Tab.~\ref{tab:chnullr028}. The performance of models trained on such datasets decreases, which indicates that the instances without entity are necessary, as they keep the distribution of the test set and training set relatively consistent. 

%In order to find what value of EnNullR of the training set and development set for the dataset can get the highest score, we make the EnNullR of the training set and development set in 0\% to 90\% with a gradient of 10\%. And we ensure that the instance numbers of these training set subsets are consistent, and so do the development set. The results are in Tab.~\ref{tab:0-1ennulr}. We find that the averaged F1 score often becomes bigger, with the EnNullR smaller. Std, however, is the opposite. WNUT16 gets the best score when EnNullR is 0.2 While Weibo is 0.1. In addition, the EnNullR of the test set in WNUT16 is 0.48 and 0.40 in Weibo.

\section{Discussion}

Our statistical evaluation framework can be used to analyze the factors that affect the dataset's quality and, furthermore, to build a higher-quality dataset in a targeted manner or augment the data with statistical improvement guidance. In this section, we take an initial step to analyze how the dataset construction process affects the statistical properties of datasets.

As shown in Tab.~\ref{tab:dataset construction}, based on an overview of the literature that presented the 10 NER datasets, we provide a summary of how they were built. We can see that all datasets were created manually, with the exception of \textit{CLUENER} and \textit{WikiAnn}. 
As for \textit{CLUENER}, \citet{xu2020cluener2020} pre-label their dataset using the distant-supervised approach with a vocabulary and then manually check and modify some labels. \textit{WikiAnn} is constructed using a cross-lingual name tagging framework based on a series of new Knowledge Base (KB) mining methods \cite{pan-etal-2017-cross}. 

We observe from Tab.~\ref{tab:r-d-v-results} that only 2 of the 10 NER datasets, \textit{CLUENER} and \textit{WikiAnn}, had Acc scores below 0.90~(90\%), indicating that the NER dataset, which was not totally created manually, will have a significant number of annotation errors (shown in Appendix Fig.~\ref{figure:mistake}).

\input{floats/dataset-construction.tex}

\section{Related Work}

\paragraph{Issues in NLP Datasets}
% From the perspective of datasets, 
Recent works have shown that NLP datasets have a number of quality problems, e.g., label mistakes~\cite{wang-etal-2019-crossweigh}, entity missing\footnote{The target summary contains entities (names, dates, events, etc.) that are absent from the source.}~\citet{Tejaswin21} in NER dataset.
%found 5.38\% label mistakes in the CoNLL03 NER dataset.
%\citet{Tejaswin21} manually checked 600 randomly selected instances from CNN/DailyMail(~\cite{hermann2015teaching};
% ~\cite{nallapati2016abstractive}), Gigaword(~\cite{rush-etal-2015-neural}), XSum(~\cite{narayan-etal-2018-dont}), and found that there is a large portion of instances belong to Entity Missing and Evidence Missing\footnote{Evidence Missing: The target summary is based on concepts which are absent from the source. However, the target is not Incomplete, and all Entities are present.} in the dataset.
In addition, researchers propose that the model performance might be inflated on some machine reading comprehension~\cite{sugawara2020assessing} datasets or natural language inference datasets~\cite{gururangan-etal-2018-annotation}. Models trained on some specific datasets reach high performance by merely relying on spurious correlations, which results in poor generalization ability.

However, most work focus on a specific issue of the datasets, and most issues are highly related to the model training process. Inspired by CTT, we build our dataset quality evaluation framework from reliability, difficulty, and validity dimensions. And we developed metrics for assessing the quality of datasets under the above three dimensions in conjunction with NER task characteristics and experimentally validated the effectiveness of our metrics in dataset evaluation.
%We XXX.

\paragraph{Data-centric AI}
In these years, the dataset is gradually being valued by the NLP and machine learning community.
\citet{Andrew21} first proposed a data-centric AI research paradigm, pointing out that focusing on optimizing data rather than models can often achieve better results. At the same time, \citet{xu2021dataclue} propose DataCLUE, the first Data-centric benchmark applied in the NLP field. The data-centric research paradigm emphasizes improving the dataset quality in an automatic or semi-automatic way to improve the model performance. We believe our proposed evaluation metrics can point the way for dataset quality improvement efforts.

% \paragraph{Fine-grained Model Performance Analysis. }
 
\section{Conclusion and Future Work}
In this paper, we investigate various statistical properties of the NER datasets and propose a comprehensive dataset evaluation framework with nine statistical metrics.
We implement a fine-grained evaluation on ten widely used NER datasets and provide a fair comparison of the existing datasets from three dimensions: difficulty, validity, and reliability. 
We further explore how the statistical properties of the training dataset influence the model performance and how do dataset construction methods affect the dataset quality.
In the future, we hope more works dive into dataset quality evaluation from a broader and more general perspective.

\end{CJK*}
\bibliographystyle{IEEEtran} 
\bibliography{anthology,custom}
\clearpage
\newpage
\appendix
\section*{Appendix}
\label{sec:appendix}

\input{floats/appendix}

\end{document}

%% file: floats/r-d-v-result.tex
\begin{table*}
\centering
\footnotesize
\setlength{\tabcolsep}{4pt}
\begin{tabular}{l rrrrrrrrr} 
\toprule[1pt]
 & \multicolumn{3}{c}{\textbf{Reliability}} & \multicolumn{4}{c}{\textbf{Difficulty}} & \multicolumn{2}{c}{\textbf{Validity}} \\	\cmidrule(lr){2-4}\cmidrule(lr){5-8}\cmidrule(lr){9-10} 				
 & \textbf{Red} \footnotesize $\downarrow$ & \textbf{Acc} \footnotesize $\uparrow$ & \textbf{LeakR} \footnotesize $\downarrow$ & \textbf{UnSeenEnR} \footnotesize $\uparrow$ & \textbf{EnAmb} \footnotesize $\uparrow$ & \textbf{EnDen} \footnotesize $\uparrow$ & \textbf{ModDiff} \footnotesize $\uparrow$ & \textbf{EnImBaD} \footnotesize $\downarrow$ & \textbf{EnNullR} \footnotesize $\downarrow$ \\ \midrule
 
\textbf{CLUENER} & \textbf{0.00} & 0.86 & \textbf{0.00} & 0.37 & 0.80 & 0.26 & \textbf{4.58} & 0.04 & \textbf{0.00} \\
\textbf{OntoNotes4} & 0.02 & 0.98 & 0.04 & 0.47 & \textbf{2.54} & \textbf{1.02} & - & 0.13 & 0.46 \\
\textbf{MSRA} & \textbf{0.00} & 0.99 & \textbf{0.00} & 0.28 & 1.16 & 0.17 & 0.38 & 0.11 & 0.41 \\
\textbf{PeopleDaily} & \textbf{0.00} & 0.96 & \textbf{0.00} & 0.22 & 1.73 & 0.65 & - & 0.11 & 0.40 \\
\textbf{Resume} & \textbf{0.00} & \textbf{1.00} & 0.01 & 0.46 & 0.29 & 0.25 & 0.41 & 0.17 & 0.17 \\
\textbf{Weibo} & 0.05 & 0.98 & 0.17 & 0.56 & 0.92 & 0.55 & 0.90 & 0.27 & 0.44 \\
\textbf{WikiAnn} & 0.03 & 0.89 & 0.13 & 0.55 & 1.53 & 0.74 & - & \textbf{0.02} & \textbf{0.00} \\ \midrule
\textbf{CoNLL03} & 0.05 & 0.96 & 0.03 & 0.46 & 0.35 & 0.28 & 0.24 & 0.06 & 0.20 \\
\textbf{WNUT16} & 0.01 & 0.97 & \textbf{0.00} & \textbf{0.89} & 0.65 & 0.51 & 2.87 & 0.08  & 0.56 \\
\textbf{OntoNotes5} & 0.01 & 0.91 & 0.03 & 0.28 & 0.76 & 0.36 & 0.64 & 0.06 & 0.55 \\

 \bottomrule[1pt]
\end{tabular}

\caption{Statistical values of different metrics under the reliability, difficulty, and validity dimensions of the 10 NER datasets (The upper rows are Chinese NER datasets, and the lower rows are English NER datasets). \textbf{\small ↑} indicates that the larger the value, the better the quality of the data set on this metric.
\textbf{\small ↓} indicates that the lower the value, the better the quality of the data set on this metric.
\textit{Red} stands for Redundancy, \textit{Acc} is Accuracy, \textit{LeakR} is Leakage Ratio, \textit{UnSeenEnR} is Unseen Entity Ratio, \textit{EnAmb} is Entity Ambiguity Degree, \textit{EnDen} is Entity Density, \textit{ModDiff} is Model Differentiation, \textit{EnImBaD} is Entity Imbalance Degree, \textit{EnNullR} is Entity-Null Rate.}
\label{tab:r-d-v-results}
\end{table*}

%% file: floats/dataset_info.tex
\begin{table}
\footnotesize
\centering

{
\begin{tabular}{c c c c} \toprule[1pt]
\textbf{Dataset} & \textbf{Lang} & \textbf{\#Tags}	& \textbf{Source} \\ \hline
CLUENER & Zh & 10 & THUCNEWS \\
OntoNotes 4 & Zh & 4 & News, Broadcast etc. \\
MSRA & Zh & 3 & News \\
PeopleDaily & Zh & 3 & News \\
Resume & Zh & 8 & Sina Finance \\
Weibo & Zh & 4 & Sina microblog \\
WikiAnn & Zh & 3 & Wikipedia \\
CoNLL03 & En & 4 & Reuters News \\
WNUT16 & En & 10 & User-generated web text \\
OntoNotes 5 & En & 18 & Broadcast etc.
\\ \bottomrule[1pt]
\end{tabular}
}
\caption{Standard Named Entity Recognition dataset statistics. Zh and En mean Chinese and English, respectively. It is important to note that OntoNotes 4 has four common tags in the Chinese dataset, although  OntoNotes 4 has a total of 18 tags (for the English dataset).}
\label{tab:dataset info}
\end{table}

%% file: floats/ennullr028.tex
\begin{table}[t]
\footnotesize
\setlength{\tabcolsep}{2pt}
\begin{tabular}{cc c c c c } 
\toprule[1pt]
    \multirow{2}{*}{\textbf{Dataset}} & \multirow{2}{*}{\textbf{Train \& Dev}} & \multicolumn{2}{c}{\textbf{W2NER}} & \multicolumn{2}{c}{\textbf{LSTM CRF}} \\\cmidrule(lr){3-4}\cmidrule(lr){5-6} 
  &  & avg.  &  std.  &  avg.  &  std.  \\ \midrule

\multirow{4}{*}{\textbf{OntoNotes5}} &  EnNullR (0.80) & 68.99 & 0.6663 & 64.80 & 0.0031 \\
 &  EnNullR (0.20) & 77.96 & 0.0134 & 75.01 & 0.0354  \\ 
 & EnNullR (0.00)  & 76.41 & 0.2037 & 74.75 & 0.0440 \\ 
 & Original & 86.99 & 0.0672 & 80.91 & 9.8596 \\ 
\hline
\multirow{4}{*}{\textbf{WNUT16}} &  EnNullR (0.80 & 49.66 & 1.3646 & 25.26 & 0.1722 \\
 & EnNullR (0.20) & 54.22 & 0.5449 & 36.99 & 0.9092 \\
 & EnNullR (0.00)  & 52.36 & 0.3577 & 36.04 & 1.3302 \\ 
 & Original  & 55.49 & 2.1829 & 36.89 & 0.0900\\  
 
 \bottomrule[1pt]
\end{tabular}
\caption{Model performance on English datasets when the proportion of samples without entities in the train set and dev set is 0.80~(80\%), 0.20~(20\%), 0.00~~(0\%), and original, respectively.}
\label{tab:ennullr028}
\end{table}

%% file: floats/chnullr028.tex
\begin{table}[t]
\centering
\footnotesize
\setlength{\tabcolsep}{2.5pt}
\begin{tabular}{cc c c c c } 
\toprule[1pt]
    \multirow{2}{*}{\textbf{Dataset}} & \multirow{2}{*}{\textbf{Train \& Dev}} &  \multicolumn{2}{c}{\textbf{Lattice LSTM}} & \multicolumn{2}{c}{\textbf{Flat-Lattice}}  \\\cmidrule(lr){3-4}\cmidrule(lr){5-6} 
  &  &avg.  &  std.  &  avg.  &  std. \\ \midrule

\multirow{4}{*}{\textbf{Weibo}} &  EnNullR (0.80) & 29.26 &5.0456& 30.96&0.3409  \\
 &     EnNullR (0.20) & 52.97 & 0.0408 & 53.55 & 2.7369 \\ 
& EnNullR (0.00)  & 54.02 & 0.3600 & 55.89 & 6.2324\\ 
 & Original  & 55.04 & 1.0192 & 57.92 & - \\ \hline
 \multirow{4}{*}{\textbf{MSRA}} &  EnNullR (0.80) & 80.51 & 0.0750 & 83.26 & 0.0157 \\
 &  EnNullR (0.20) & 91.11 & 0.0151
 & 92.82 & 0.0259  \\ 
& EnNullR(0.00)  & 91.94 & 0.0001 & 93.60 & 0.0097\\ 
 & Original  & 92.50 & 0.0273 & 94.06  & - \\ 
 \bottomrule[1pt]
\end{tabular}
\caption{Model performance on Chinese datasets when the proportion of samples without entities in the train set and dev set is 0.80~(80\%), 0.20~(20\%), 0.00~(0\%), and original, respectively.}
\label{tab:chnullr028}
\end{table}

%% file: floats/dataset-construction.tex
\begin{table}
\small
\setlength{\tabcolsep}{2pt}
\centering
{
\begin{tabular}{c c} \toprule[1pt]
\textbf{Dataset} & \textbf{Construction Method} \\ \hline
CLUENER & distant supervision + human\\
OntoNotes4 & human annotation  \\
MSRA & human annotation \\
PeopleDaily & human annotation\\
Resume  & human annotation\\
Weibo  & human annotation\\
WikiAnn  &  cross-lingual
name tagging framework \\
CoNLL03  & human annotation\\
WNUT16  & human annotation\\
OntoNotes5  & human annotation\\
\bottomrule[1pt]
\end{tabular}
}
\caption{Standard Named Entity Recognition dataset construction method. }
\label{tab:dataset construction}
\end{table}

%% file: floats/appendix.tex
\section{Settings for Different Models}\label{app:setting}
W2NER, Roberta, and LUKE adopt AdamW \cite{loshchilov2017decoupled} optimizer. Stochastic Gradient Descent (SGD) is used for optimation in Lattice LSTM, LSTM CRF, and Flat-Lattice. 
Furthermore, LUKE and W2NER optimize the model using AdamW with learning rate warmup and linear decay of the learning rate. LUKE also uses early stopping based on performance on the development set. The hyper-parameters of six models are shown below. In addition, the hyper-parameter used for the large dataset is the left one in the bracket.

\section{Word Embeddings}\label{app:emb}
\paragraph{Static Word Embeddings}
Lattice LSTM uses its own word~\footnote{https://github.com/jiesutd/RichWordSegmentor}, character and character bigram embeddings~\footnote{https://github.com/jiesutd/LatticeLSTM}. 
%Character embeddings are pretrained on Chinese Giga-Word~\footnote{https://catalog.ldc.upenn.edu/LDC2011T13} using word2vec~(~\cite{mikolov2013efficient}) and fine-tuned at model training.
LSTM CRF also has its own pretained embedding, but it is not available. So we use the embedding of common-crawl vectors from 
fasttext~\footnote{https://fasttext.cc/docs/en/english-vectors.html} instead.
Flat-Lattice uses the same pretained embeddings as Lattice-LSTM. 
\paragraph{Dynamic Word Embeddings}
Words have different meanings in different contexts, while static word embedding cannot consider different contexts. The word embedding representation obtained through BERT integrates more grammatical, lexical, and semantic information. Besides, dynamic word embedding can also enable words to have different word embedding representations in different contexts. LUKE proposes new pretrained contextualized representations of words and entities with Roberta. W2NER use bert-large-cased for the English dataset and bert-base-chinese for Chinese. Roberta is undoubtedly an optimized version of BERT.
\input{floats/hparam_w2ner.tex}
\input{floats/hparam_BiLSTM.tex}

\input{floats/hparam_Flat_big.tex}

\input{floats/hparam_roberta.tex}
\input{floats/hparam_Lattice.tex}
\input{floats/hparam_luke.tex}

\section{Validation Details of the Metrics under our Statistical Evaluation Framework}
We justify and clarify those metrics under our evaluation framework that we have not discussed further in the main text.

\paragraph{Redundancy} International standards for data \footnote{\url{https://iso25000.com/index.php/en/iso-25000-standards/iso-25012}} require the uniqueness of data. Data sets are a special data type and should also follow the corresponding standards.

\paragraph{Accuracy} Numerous research have demonstrated that flaws in the data set will negatively impact the model's performance \cite{zhu2003eliminating,Tejaswin21,gupta2019dealing}. The model's performance will increase to some extent after these mistakes are fixed \cite{zeng2021validating}.

\paragraph{Text Complexity} Several experiments of \citet{fu2020interpretable} on the English NER datasets support our use of entity density as a valid metric of the difficulty of the dataset. Their experiments showed that NER models are negatively correlated with entity density.

\paragraph{Model Differentiation} This metric is an extrinsic metric that aims to assess the dispersion of model scores on a unified benchmark dataset. As long as enough models are evaluated on the data set, we can measure the differentiation of a dataset by calculating the dispersion of the scores of different models. 

\paragraph{Entity Imbalance Degree}
There are category imbalances in many NLP tasks that can seriously affect the model's performance on the long-tail instances \cite{blevins2020moving,zhang2021deep,wang2020infobert}. Therefore, the Entity Imbalance Degree of the NER dataset is necessary and practical.

\paragraph{Leakage Ratio}
% \paragraph{The model performs better on the instances it has seen during training.}

We validate the metric following the controlled data adjustment method in 
 Sec. ~\ref{sec:validate}.
As shown in Tab.~\ref{tab:LeakageExp}, 3 models (i.e., Lattice LSTM, Flat-Lattice, and Roberta) consistently achieve better performance when the leakage rate of the test set is 0.80~(80\%) than when it is 0.20~(20\%). In particular, we found that the performance of  Flat-Lattice on the Weibo test set with a Leakage Ratio of 0.80~(80\%) outperformed the 0.20~(20\%) by a large margin i.e., 25.69\%. We speculate that because the model has seen the leaked data in the test set during training, it performs better on the test set with a relatively high data leakage rate. Looking at the experimental results from another perspective, researchers need to pay more attention to how to improve the generalization ability of the NER model.

\input{floats/28weibo_leakageRatio.tex}

\begin{figure}[t]
    %\centering
    \centerline{\includegraphics[width=1\linewidth]{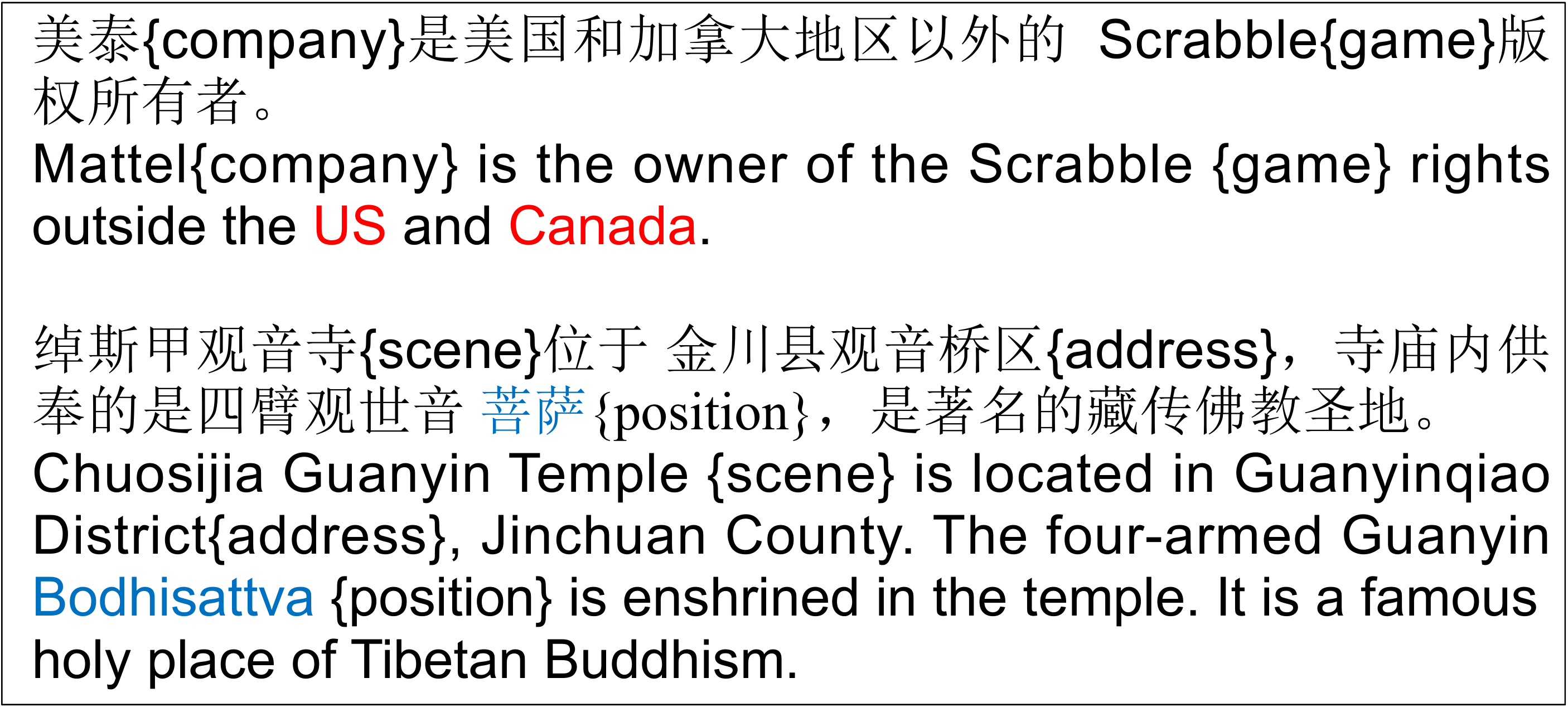}}
    \caption{Mislabeled Examples of randomly selected samples from \textit{CLUNER}. \textcolor{red}{red} indicates missing entities not assigned entity labels. \textcolor{blue}{blue} indicates the entity with the wrong labeled entity type.}
    \label{figure:mistake}
    \vspace{-0.2cm} %图片与正文间距
\end{figure}

\section{Mistakes in NER Datasets}\label{app:annotation accuracy}

As stated in Sec.~\ref{formu}, we selected 100 instances from
each dataset split and invited at least three professional linguists to annotate the accuracy. Fig.~\ref{figure:mistake} presents some samples of entities that are consistently considered to be incorrectly labeled in \textbf{CLUNER}.

\section{Specific assessment results for different splits of the dataset under the reliability metrics}\label{appendix:Model Replication}
As introduced in Sec.~\ref{sec:42}, we average the scores of the training, the development, and the test split of the datasets for Redundancy and Accuracy to obtain Redundancy and Accuracy of the corresponding datasets. 
Tab.~\ref{tab:ENER-Reliability Results} presents evaluated split values for different English NER datasets, and Tab.~\ref{tab:16} is for Chinese NER datasets.

\input{floats/ener_results_reliability.tex}
\input{floats/cner_results_reliability.tex}

\section{Model Replication Results}\label{app:Model Replication}
In the paper, we have reproduced \textbf{Lattice LSTM}, \textbf{Flat-Lattice}, \textbf{Roberta}, \textbf{LSTM CRF}, \textbf{LUKE}, and \textbf{W2NER}. The results of our reproduced models are compared to the performance of the cited works are listed in Tab.~\ref{tab:13} and Tab.~\ref{tab:14}.

\input{floats/model_reproduction_cner.tex}
\input{floats/model_reproduction_ener.tex}

%% file: floats/hparam_w2ner.tex
\begin{table}[]
  \begin{center}
    
    \begin{tabular}{l c}
      \hline
      \textbf{Hyper-parameter} & \textbf{Value}
      \\
      \hline
       dist\_emb\_size & 20 \\
      type\_emb\_size & 20  \\
      bert\_hid\_size & [768, 1024]  \\
      conv\_hid\_size & [96, 64]\\
      lstm\_hid\_size & [768, 512]\\
      dropout    &  0.5        \\
      learning rate (BERT)& [1e-5, 5e-6]\\
      learning rate (others) & 1e-3\\
     batch size & [2, 4, 8]\\
      \hline
    \end{tabular}
    \caption{Hyper-parameter settings on the W2NER.}
    \label{hyper-w2ner}
  \end{center}
\end{table}

%% file: floats/hparam_BiLSTM.tex
\begin{table}[]
  \begin{center}
    
    \begin{tabular}{l c}
      \hline
      \textbf{Hyper-parameter} & \textbf{Value}
      \\
      \hline
      gradient clipping & 5.0\\
      layer dimension & 100\\
       LSTM layer & 1\\   
      dropout    &  0.5        \\
     char\_dim & 25\\
     char\_lstm\_dim & 25\\
     word\_dim & 300\\
     word\_lstm\_dim & 100\\
      learning rate & 0.01\\
    
      \hline
    \end{tabular}
    \caption{Hyper-parameter settings on the LSTM CRF.}
    \label{tab:hyper-bilstmcrf}
  \end{center}
\end{table}

%% file: floats/hparam_Flat_big.tex
\begin{table}[t]
  \begin{center}
    
    \begin{tabular}{l c }
      \hline
      \textbf{Hyper-parameter} & \textbf{Value}
      \\
      \hline
       decay & -0.05 \\
      momentum & -0.9  \\
      FFN\_size & 480\\
      head & [8, 4, 12]  \\
      d\_{head} & [16,20]  \\
      d\_{model} & head $\times$ d\_{head} \\
      embed dropout    &  0.5        \\
      output dropout    &  0.3        \\
      learning rate & [1e-3, 8e-4]\\
      warmup & [10, 1, 5](epoch)  \\
     batch size & [10, 8]\\
      \hline
    \end{tabular}
    \caption{Hyper-parameter settings on the Flat-Lattice.}
  \end{center}
\end{table}

%% file: floats/hparam_roberta.tex
\begin{table}[t]
  \renewcommand\arraystretch{1.2}
  \begin{center}
    
    \begin{tabular}{l c}
      \hline
      \textbf{Hyper-parameter} & \textbf{Value}
      \\
      \hline
      embedding size & 50\\
      LSTM hidden & 200\\
       batch size & 1\\ 
       learning rate &0.015\\
      dropout    &  0.5        \\    
      learning rate & 5e-5\\
    
      \hline
    \end{tabular}
    \caption{Hyper-parameter settings on the Lattice LSTM.}
  \end{center}
\end{table}

%% file: floats/hparam_Lattice.tex
\begin{table}[t]
  \begin{center}
    \begin{tabular}{l c}
      \hline
      \textbf{Hyper-parameter} & \textbf{Value}
      \\
      \hline
     batch size & 32 \\
     max sentence length & 300\\
     weight decay rate & 0.1  \\
            warmup & 100(step) \\
      \hline
    \end{tabular}
    \caption{Hyper-parameter settings on the Roberta.}
  \end{center}
\end{table}

%% file: floats/hparam_luke.tex
\begin{table}[t]
  \begin{center}
    
    \begin{tabular}{l c}
      \hline
      \textbf{Hyper-parameter} & \textbf{Value}
      \\
      \hline
      batch size & [4, 8]\\
        adam $\beta_1$  & 0.9\\
        adam $\beta_2$  & 0.98\\
        adam $\epsilon$  & 1e-6\\
      dropout    &  0.1       \\
     warmup ratio & 0.06\\
     weight decay & 0.01\\
     maximum word length & 512\\
     %word$\_$lstm$\_$dim & 100\\
      learning rate & 1e-5\\
    gradient clipping & none \\
      \hline
    \end{tabular}
    \caption{Hyper-parameter settings on LUKE.}
    \label{hyper-luke}
  \end{center}
\end{table}

%% file: floats/28weibo_leakageRatio.tex
\begin{table}[htbp]
\footnotesize
\setlength{\tabcolsep}{3pt}
\begin{tabular}{c c c c c}
\toprule[1pt]
\textbf{Dataset} & \textbf{Test} & \textbf{LSTM} & \textbf{Flat-Lattice} & \textbf{Roberta} \\
% \cmidrule(lr){1-}\cmidrule(lr){2-}\cmidrule(lr){3-5}
\midrule
\multirow{2}{*}{\textbf{Weibo}} & Leakage (80\%) & \textbf{73.68} & \textbf{74.24} & \textbf{78.52}   \\ 
& Leakage (20\%) & 64.86 & 48.65 & 70.40  \\ \bottomrule[1pt]
\end{tabular}

\caption{Model performance when the proportion of leaked samples in the test set is 80\% and 20\%, respectively. LSTM represents Lattice LSTM.}
\label{tab:LeakageExp}
\end{table}

%% file: floats/ener_results_reliability.tex
\begin{table}[t]
\small
\setlength{\tabcolsep}{3pt}
\begin{tabular}{c ccc ccc cccc} \toprule[1pt]
  &  \multicolumn{3}{c}{\textbf{CoNLL03}} & \multicolumn{3}{c}{\textbf{WNUT16}}  & \multicolumn{3}{c}{\textbf{OntoNotes5}}\\ \cmidrule(lr){2-4}\cmidrule(lr){5-7}\cmidrule(lr){8-10}
  &  \multicolumn{1}{c}{train} & dev & test & train & dev & test & train & dev & test \\ \midrule
 
 \textbf{Red} & \multicolumn{1}{c}{0.06}	 & 0.03 & 	0.05 & 	0.04 & 0.00 & 0.00 & 0.01 & 	0.01 & 	0.01 \\ 
  \textbf{Acc} & \multicolumn{1}{c}{0.93}	 & 0.96 & 0.98 & 0.95 & 0.98 & 0.97 & 0.90	& 0.88	 & 0.95 \\ 
  %\textbf{LeakR} & \multicolumn{3}{c}{0.03} & \multicolumn{3}{c}{0.00} & \multicolumn{3}{c}{0.03}\\ 
  \bottomrule[1pt]
\end{tabular}
\caption{Results of metrics (except Leakage Ratio) under the reliability dimension of English named entity recognition datasets. Red and Acc denote Redundancy and  Accuracy, respectively.}
\label{tab:ENER-Reliability Results}
\end{table}

%% file: floats/cner_results_reliability.tex
\begin{comment}
\begin{table*}[b]
\centering
\resizebox{15.5cm}{!}{
\begin{tabular}{ccc ccc cc ccc ccc ccc ccc} \toprule[1pt]
&  \multicolumn{2}{c}{\textbf{ClUENER}} & \multicolumn{3}{c}{\textbf{OntoNotes4}} & \multicolumn{2}{c}{\textbf{MSRA}} & \multicolumn{3}{c}{\textbf{PeopleDaily}} & \multicolumn{3}{c}{\textbf{Resume}}  & \multicolumn{3}{c}{\textbf{Weibo}}   & \multicolumn{3}{c}{\textbf{WikiAnn}} \\ \cmidrule(lr){2-3}\cmidrule(lr){4-6}\cmidrule(lr){7-8}\cmidrule(lr){9-11}\cmidrule(lr){12-14}\cmidrule(lr){15-17}\cmidrule(lr){18-20}
 & \multicolumn{1}{c}{train} & dev & train & dev & test & train & test & train & dev & test & train & dev & test & train & dev & test & train & dev & test  \\  \midrule
 
 \textbf{Red} & \multicolumn{1}{c}{0.00} & 0.00 & 0.02 & 0.03 & 0.01 & 0.00 & 0.00 & 0.00 & 0.00 & 0.00 & 0.00 & 0.00 & 0.00 & 0.08 & 0.03 & 0.03 & 0.04 & 0.03 & 0.03 \\ 
 \textbf{Acc} & \multicolumn{1}{c}{0.89} & 0.83 & 0.98 & 0.97 & 0.98 & 0.99 & 1.00 & 0.96 & 0.94 & 0.97 & 1.00 & 1.00 & 1.00 & 0.96 & 0.98 & 0.99 & 0.90 & 0.88 & 0.90 \\ 
 \bottomrule[1pt]
\end{tabular}
}
\caption{\label{tab:CNER-Reliability Results}
Results of metrics under the reliability dimension of Chinese named entity recognition datasets. Red, Acc, and LeakR denote Redundancy, Accuracy, and Leakage Ratio, respectively.}
\end{table*}
\end{comment}

\begin{table}[H]
\small
\setlength{\tabcolsep}{15pt}
 \begin{tabular}{cccc} \toprule[1pt]
 \textbf{Dataset} & \textbf{Split} & \textbf{Red} & \textbf{Acc} \\ \midrule
 \multirow{2}{*}{\textbf{CLUENER}} & train &  0.00 & 0.89 \\
  & dev & 0.00 & 0.83 \\ \midrule
 \multirow{3}{*}{\textbf{OntoNotes4}} & train & 0.02 & 0.98 \\
  & dev & 0.03 & 0.97 \\ 
  & test & 0.01 & 0.98 \\ \midrule
 \multirow{2}{*}{\textbf{MSRA}} & train & 0.00 & 0.99 \\
  & test & 0.00 & 1.00  \\ \midrule
 \multirow{3}{*}{\textbf{PeopleDaily}} & train & 0.00 & 0.96\\
  & dev & 0.00 & 0.94 \\
  & test & 0.00 & 0.97 \\ \midrule
 \multirow{3}{*}{\textbf{Resume}} & train & 0.00 & 1.00 \\
  & dev & 0.00 & 1.00 \\
  & test & 0.00 & 1.00 \\ \midrule
 \multirow{3}{*}{\textbf{Weibo}} & train & 0.08 & 0.96 \\
  & dev & 0.03& 0.98 \\
  & test & 0.03 & 0.99 \\ \midrule
 \multirow{3}{*}{\textbf{WikiAnn}} & train & 0.04 & 0.90 \\
  & dev & 0.03 & 0.88 \\
  & test & 0.03 & 0.90 \\
 \bottomrule[1pt]
 \end{tabular}
\caption{\label{tab:16}
Results of metrics (except Leakage Ratio) under the reliability dimension of Chinese named entity recognition datasets. Red and Acc denote Redundancy and Accuracy, respectively.}
\end{table}

%% file: floats/model_reproduction_cner.tex
\begin{table}[H]
\small
\setlength{\tabcolsep}{4pt}
\begin{tabular}{l c c c c c c} \toprule
  &  \multicolumn{2}{c}{\textbf{Lattice LSTM}} & \multicolumn{2}{c}{\textbf{Flat-Lattice}}  & \multicolumn{2}{c}{\textbf{Roberta}}\\ \cmidrule(lr){2-3}\cmidrule(lr){4-5}\cmidrule(lr){6-7}
  
  &  ori.  &  repro.  &  ori.  &  repro.  &  ori.  &  repro.\\ \midrule
MSRA & 93.18 & 93.12 & 94.35 & 94.06 & - & 94.57 \\ 
OntoNotes4  & 73.88 & 73.43 & 75.70 & 75.84 & - & 80.30 \\ 
Resume & 94.46 & 94.46 & 94.93 & 95.11 & - & 96.19 \\
Weibo & 58.79  & 56.49  & 63.42  & 57.92  & - & 67.92  \\ \bottomrule
\end{tabular}
\caption{repro. denotes reproduction - denotes that the authors of the literature we cited did not experiment on that dataset. And ori. denotes original paper results.}
\label{tab:13}
\end{table}

%% file: floats/model_reproduction_ener.tex
\begin{table}[H]
\centering
\small
\setlength{\tabcolsep}{2pt}
\begin{tabular}{c c c c c c c} \toprule
  &  \multicolumn{2}{c }{\textbf{LSTM CRF}} & \multicolumn{2}{c }{\textbf{W2NER}}  & \multicolumn{2}{c}{\textbf{LUKE}}\\ \cmidrule(lr){2-3}\cmidrule(lr){4-5}\cmidrule(lr){6-7}
  
  &  ori.  &  repro.  &  ori.  &  repro.  &  ori.  &  repro.\\ \midrule
CoNLL03 & 83.63 & 83.61 & 93.07 & 92.02 & 94.30 & 94.2 \\ 
WNUT16  & - & 26.04 & - & 45.81 & - & 56.99 \\ 
OntoNotes5 & - & 80.14 & 90.50 & 84.92 & - & 87.27 \\ \bottomrule
\end{tabular}
\caption{repro. denotes reproduction - denotes that the authors of the literature we cited did not experiment on that dataset. And ori. denotes the original paper results }
\label{tab:14}
\end{table}